\documentclass{article} 
\usepackage{iclr2027_conference,times}

\iclrfinalcopy

\usepackage{amsmath,amsfonts,bm}

\def\eqref#1{equation~\ref{#1}}

\def\1{\bm{1}}

\DeclareMathAlphabet{\mathsfit}{\encodingdefault}{\sfdefault}{m}{sl}
\SetMathAlphabet{\mathsfit}{bold}{\encodingdefault}{\sfdefault}{bx}{n}

\usepackage{url}

\definecolor{cvprblue}{rgb}{0.21,0.49,0.74}

\usepackage[pagebackref,breaklinks,colorlinks,allcolors=cvprblue]{hyperref}

\usepackage{enumitem}
\usepackage{booktabs}
\usepackage{array}
\usepackage{multirow}
\usepackage{graphicx}
\usepackage{colortbl}
\usepackage[table]{xcolor}

\usepackage{algorithm}
\usepackage{algpseudocode}
\usepackage{amsmath}
\usepackage{amssymb}
\usepackage{amsthm}

\definecolor{myblue}{RGB}{45,120,175}
\definecolor{myred}{RGB}{205,70,75}

\title{Ego4WAM: What Matters When Scaling Egocentric Human Data for Robot Learning?}

\author{
Zhihao Sun$^{1}$\quad
Liu Liu$^{2\dagger}$\quad
Xinjiang Wang$^{2}$\quad
Haoyi Jiang$^{3}$\quad
Wei Feng$^{2}$
\\
\textbf{
Huiqiang Zhang$^{4}$\quad
Xiaosong Jia$^{1}$\quad
Zhizhong Su$^{2}$\quad
Zuxuan Wu$^{1\ddagger}$
}\\[3pt]
$^{1}$Institute of Trustworthy Embodied AI, Fudan University \quad
$^{2}$Horizon Robotics\\
$^{3}$Huazhong University of Science \& Technology \quad
$^{4}$Zhejiang University of Technology\\
$^{\dagger}$Project Leader \quad $^{\ddagger}$Correspondence Author\\[2pt]
\texttt{sunzhihao.18@gmail.com, zxwu@fudan.edu.cn}
}

\begin{document}

\maketitle

\begin{abstract}
Egocentric human data provides a scalable source of experience for robot learning, but varies substantially in human-robot alignment, behavioral coverage, and available supervision. Existing work shows favorable scaling with increasing human data, but it remains unclear which data properties drive downstream robot gains and how to use such data throughout the training pipeline. We present a systematic study of egocentric human data with different alignment and supervision under a unified world-action model framework. With the model backbone fixed, we disentangle the effects of human-robot alignment, data duration and task diversity, action supervision, and data usage strategies. We find that aligned human demonstrations substantially improve out-of-distribution generalization and reduce target-task robot data requirements; data duration and task diversity affect downstream capabilities differently; and video-only supervision remains effective without action labels, providing a strong foundation for subsequent video-action training. We validate these findings through closed-loop policy evaluation on both real robots and RoboDojo. Rather than treating data duration as the sole scaling axis, Ego4WAM shows how alignment, task diversity, available supervision, and usage strategy jointly shape the value of egocentric human data for robot learning. The project page is available at
{\hypersetup{urlcolor=magenta}\href{https://sunzhihao18.github.io/Ego4WAM/}{sunzhihao18.github.io/Ego4WAM}}.
\end{abstract}

\section{Introduction}
\label{sec:introduction}

Egocentric human data offers a scalable source of experience for robot learning, covering a breadth of objects, scenes, and task variations that is difficult to match through robot teleoperation alone~\citep{Ego4D,EpicKitchens100,EgoDex}. Such experience can provide different forms of supervision for robot learning. Human hand actions provide action supervision for imitation learning~\citep{PiEgo,VITRA}, while World-Action Models (WAMs)~\citep{EgoWAM} additionally leverage future-state prediction as world-modeling supervision to learn task-relevant dynamics. Recent studies have shown strong human-to-robot transfer and favorable scaling with increasing egocentric data~\citep{Pi05,EgoScale,BeingH07,Dyna2}, yet it remains unclear which data properties drive downstream robot gains and how to use the data throughout the training pipeline. This raises a central question: \textbf{When scaling egocentric human data for robot learning, which data properties and usage strategies matter most?}

Egocentric data varies along several dimensions that directly affect its utility for robot learning. At one end of the spectrum, human demonstrations can be carefully aligned with robot data in viewpoint, motion speed, and behavior style, facilitating cross-embodiment learning but limiting the diversity of experience that can be collected. Relaxing these alignment constraints allows egocentric datasets to cover a much wider range of tasks, objects, and environments. However, obtaining reliable human action trajectories requires accurate camera-pose and hand-pose estimation, introducing additional processing cost and reducing label reliability~\citep{ACEEgo0,EgoScale}. Nevertheless, videos without action labels still contain informative interaction dynamics that WAMs can exploit through world modeling~\citep{Egomimic,EgoWAM,Dyna2}. These factors are often entangled when egocentric data is summarized only by duration, making it difficult to determine whether downstream gains arise from stronger human-robot alignment, broader behavioral coverage, different training objectives, or increased data scale itself.

To disentangle these factors, we conduct a systematic study of how to scale and use egocentric human data for robot learning under a unified WAM framework. We keep the model backbone fixed across our main experiments, while varying data alignment, coverage, supervision, and usage strategy. We first study human-robot alignment by examining whether aligned human demonstrations can improve out-of-distribution robot generalization and reduce the need for robot data. We then separate data duration from task diversity, comparing denser coverage of existing tasks with broader coverage across different behaviors. Finally, we relax action supervision and investigate whether large-scale egocentric videos without reliable action labels can still contribute through world modeling, and how such supervision interacts with subsequent video-action training. In this way, we disentangle data properties and usage strategies that aggregate scaling otherwise conflates.

Crucially, we evaluate the utility of egocentric data by measuring its effect on closed-loop policy performance on both real-world and RoboDojo~\citep{RoboDojo} manipulation tasks, rather than relying on human-action prediction MSE alone. Across these studies, we find that egocentric data provides distinct benefits under different conditions. Human demonstrations covering object and scene conditions absent from robot training substantially improve OOD generalization and reduce the amount of target-task robot data required. Increasing data duration and task diversity leads to different performance trends across downstream capabilities. Video-only experience remains valuable beyond the subset with reliable action supervision, providing a strong foundation for subsequent video-action training. These findings show that the value of egocentric data depends not on scale alone, but on the distributions it covers, the supervision it reliably provides, and how it is used alongside robot data across the training pipeline.
Our contributions can be summarized as follows:

\begin{itemize}[leftmargin=*]
    \item We study the utility of egocentric human data for robot learning under a controlled WAM framework. With the WAM backbone fixed, we disentangle data properties and usage strategies that large-scale scaling studies often couple.

    \item We characterize how human-robot alignment, data duration and task diversity, and available supervision affect downstream robot learning. This reveals distinct roles of these factors that are obscured when egocentric data is characterized only by its overall scale.

    \item We evaluate these effects directly through closed-loop policy performance on real robots and simulation rather than relying on human-action prediction metrics alone. Our results provide practical guidance on what egocentric data to collect, what supervision to obtain, and how to use human data across robot learning.
\end{itemize}

\section{Related Works}
\label{sec:related_work}

\begin{figure}[t]
\begin{center}
\includegraphics[width=0.95\linewidth]{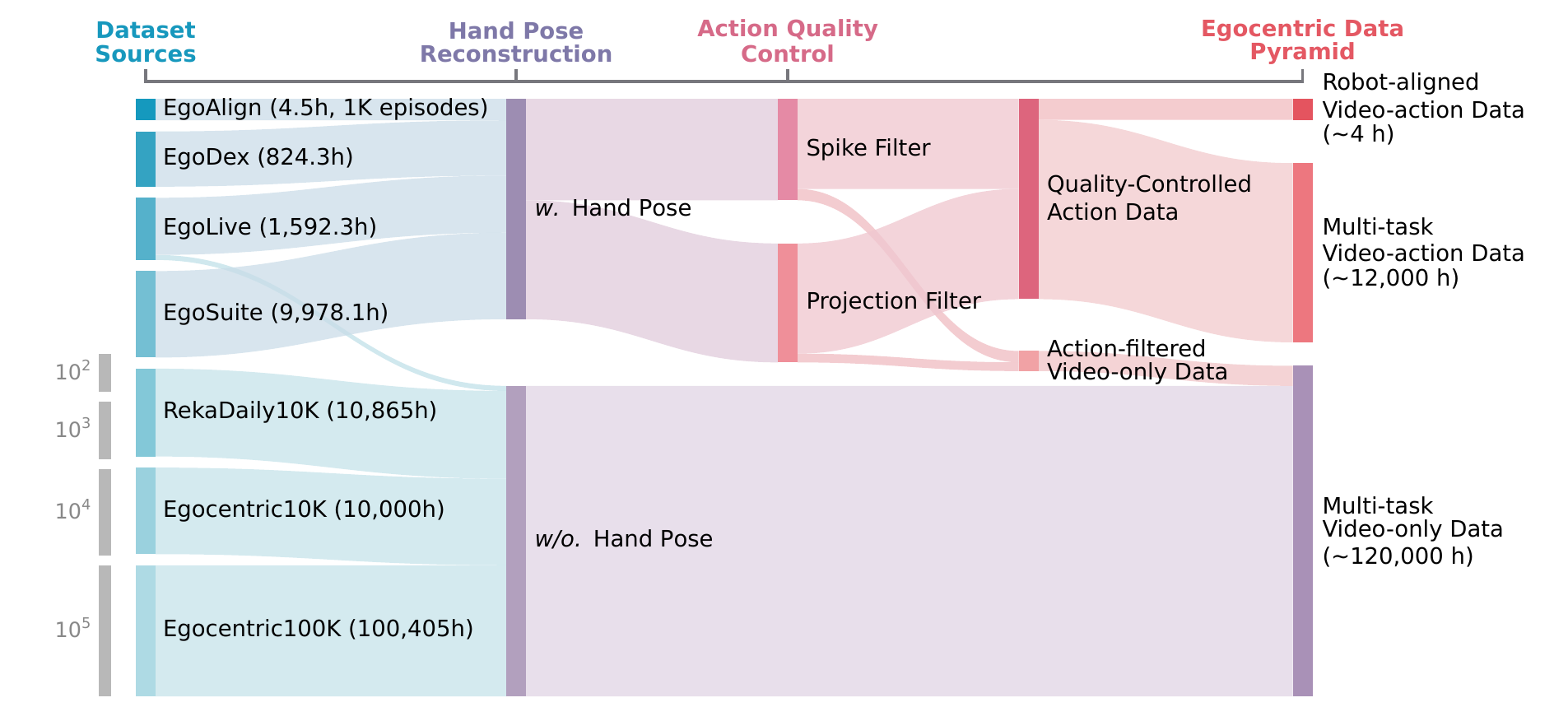}
\end{center}
\vspace{-1em}
\caption{Overview of the egocentric data pyramid.}
\label{figure:egocentric_data_pyramid}
\end{figure}

\vspace{0.05in}
\noindent\textbf{Egocentric Human Data.}
Egocentric human data has attracted increasing attention in robot learning because it can be collected at substantially lower cost than robot teleoperation while covering a much broader range of objects, scenes, tasks, and interaction patterns.
Large-scale datasets such as Ego4D~\citep{Ego4D}, Egocentric-10K~\citep{Egocentric10K}, and RekaDaily-10K~\citep{RekaDaily10K} primarily capture first-person activities from everyday life and work, providing large-scale visual and behavioral experience.
More recent manipulation-oriented datasets, including EgoDex~\citep{EgoDex}, EgoLive~\citep{EgoLive}, and EgoSuite~\citep{EgoSuite}, further augment egocentric observations with structured hand pose annotations~\citep{HaMeR}.
Such action annotations make human experience more directly usable for robot policy learning, but obtaining reliable action trajectories introduces additional sensing, reconstruction, and quality-control requirements, limiting the amount of data that can be annotated with high confidence.
A further step is to collect human demonstrations that are better aligned with downstream robot tasks. For example, EgoVerse~\citep{EgoVerse} considers task-aligned and environment-aligned human data for human-to-robot transfer. Although stronger alignment facilitates transfer to robot learning, it also imposes stricter collection requirements.
Consequently, modern egocentric data naturally spans a spectrum from large-scale video-only experience to action-annotated data and robot-aligned demonstrations, with different levels of alignment and supervision.
As illustrated in Figure~\ref{figure:egocentric_data_pyramid}, we organize this spectrum as an \emph{egocentric data pyramid}, retaining each sample at the most reliable supervision level it can support. This perspective provides a basis for studying which data properties deserve more attention, what supervision is worth obtaining, and how to use egocentric data as its scale continues to grow.

\vspace{0.05in}
\noindent\textbf{Utilizing Egocentric Data for Robot Learning.}
Recent work increasingly studies how egocentric human experience can support robot learning across different training settings. For task-aligned transfer, EgoVerse~\citep{EgoVerse} shows that broad human experience can be adapted effectively with a small amount of aligned human--robot data, while EgoWAM~\citep{EgoWAM} investigates how different world-prediction objectives transfer human experience under controlled human--robot alignment.
At larger scale, Being-H0~\citep{BeingH0} and EgoScale~\citep{EgoScale} use retargeted human actions to pretrain transferable manipulation priors, while Being-H0.7~\citep{BeingH07} and OpenWAM~\citep{OpenWAM} further introduce future or video prediction so that egocentric observations can contribute through world-model supervision. Complementary efforts reduce the embodiment gap by explicitly transforming human experience toward the robot domain through hand and wrist retargeting, action-space or viewpoint alignment, and visual embodiment editing~\citep{Ego2Robot}. Such conversion can make human data easier to incorporate into robot policies, but also introduces additional dependence on camera tracking, hand reconstruction, and visual synthesis, whose errors and embodiment ambiguities can limit reliability.
More recently, EgoScale~\citep{EgoScale}, Motus2~\citep{Motus2}, and Dyna-2~\citep{Dyna2} have demonstrated favorable scaling trends with increasingly large human-data mixtures. These results establish the scalability of human experience, but aggregate scaling curves and open-loop human prediction metrics do not fully reveal which properties of egocentric data drive downstream robot gains, or whether their relative benefits persist under closed-loop deployment. Rather than focusing on converting human data into robot-compatible supervision, we systematically study data properties and usage strategies under a controlled WAM framework. We evaluate these factors directly through closed-loop robot performance, aiming to provide actionable guidance on what egocentric data to collect, what supervision to retain, and how to use such data for robot learning.
\section{Method}
\label{sec:method}

Our goal is to provide a controlled framework for studying how egocentric data contributes to robot learning under different supervision and alignment settings. We therefore keep the model backbone and unified action space fixed across our main experiments, while varying the data composition, supervision, and usage strategies.

\subsection{Egocentric Data Pyramid}

As discussed in Section~\ref{sec:related_work} and illustrated in Figure~\ref{figure:egocentric_data_pyramid}, egocentric data varies substantially in human-robot alignment and available supervision. We organize these differences into an egocentric human data pyramid, ranging from a small amount of robot-aligned human demonstrations to large-scale action-annotated data and a larger collection of video-only data.

EgoAlign is our internally collected dataset, processed separately to form the robot-aligned subset with closely matched task and behavioral distributions. We obtain human actions through a hand-pose and camera-pose annotation pipeline, with implementation details in the appendix. For public datasets with available action annotations, we further apply temporal and geometric filters to control action quality, with the detailed filtering criteria described in the appendix. Samples that pass these filters form our quality-controlled multi-task video-action data.
We retain samples without action annotations, along with those rejected by the action-quality pipeline, as video-only data whenever the visual interaction remains valid. Although they do not provide reliable action supervision, they still capture object motion, interaction outcomes, and scene transitions that can support world modeling. Together, these data constitute the egocentric data collection from which we construct the training subsets used in our subsequent experiments.

\subsection{World-Action Model Framework}
\label{sec:wam_framework}

\begin{figure}[t]
\begin{center}
\includegraphics[width=\linewidth]{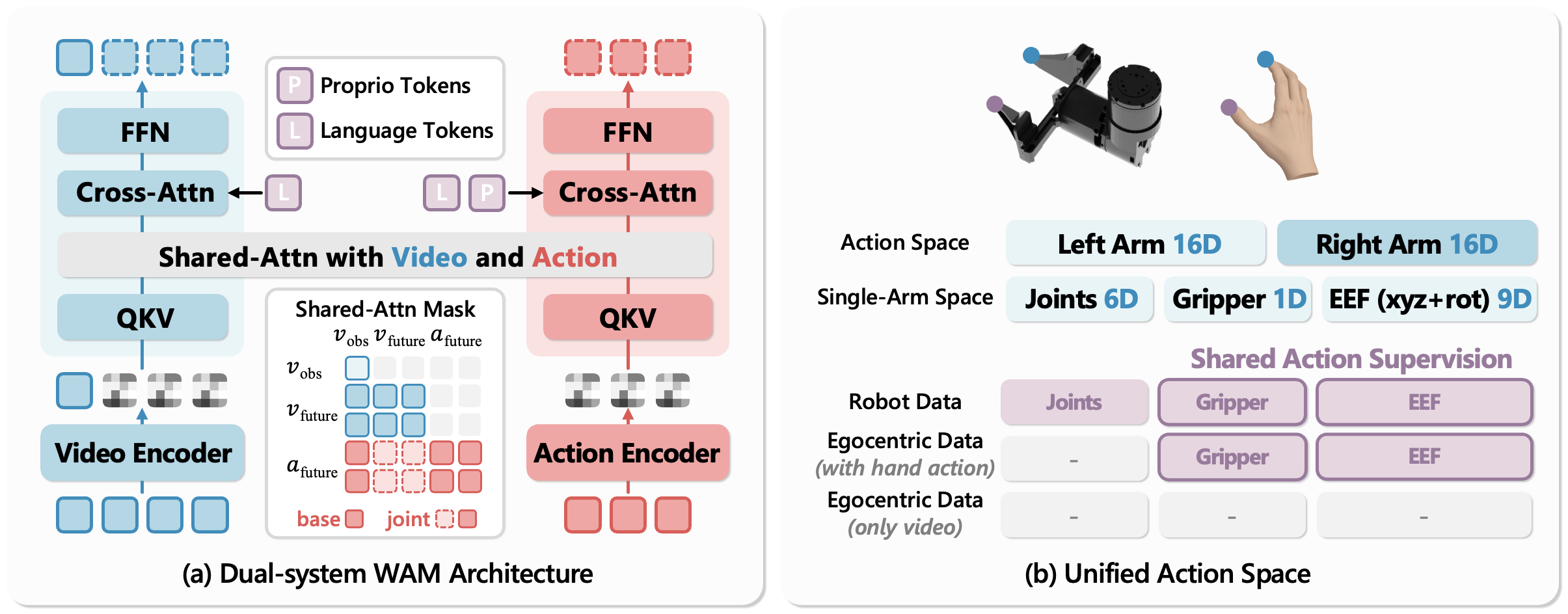}
\end{center}
\vspace{-1em}
\caption{Illustration of the world-action model framework and unified human-robot action space.}
\label{figure:method}
\end{figure}

Building on~\cite{FastWAM}, we design the backbone used throughout our study, as illustrated in Figure~\ref{figure:method} (a). The model consists of a video branch for world modeling and an action branch for action prediction. Language instructions are used as conditioning signals, while proprioceptive states are provided only to the action branch and are not injected into the world-modeling branch. The two branches interact through shared attention. We consider two shared-attention mechanisms in our study. In the \emph{base} formulation, future action tokens attend only to the current visual context. In the \emph{joint} formulation, they can additionally attend to predicted future visual representations, while future visual tokens remain independent of action tokens in both settings. This formulation naturally supports different supervision regimes. Video-action data can jointly supervise world and action prediction, while video-only data can independently supervise world modeling.

\subsection{Unified Action Space}

Human and robot actions differ substantially in their embodiment-specific control dimensions. Rather than fully retargeting human motion to robot joint commands, we define a unified action space that explicitly separates robot-specific control from components shared across embodiments. As shown in Figure~\ref{figure:method} (b), all tasks in our study use bimanual control. Each arm is represented by a 16D action vector consisting of a 6D joint signal, a 1D gripper state, and a 9D end-effector (EEF) representation, where the EEF component contains 3D translation and a 6D rotation representation. We concatenate the two arms' actions into a 32D bimanual action space.

For egocentric human data with action annotations, we supervise only the components that admit a meaningful cross-embodiment correspondence. Following prior human-robot co-training practice, we use only the EEF pose and gripper state as the default shared representation for human motion. To handle the missing dimensions in the flow-matching action space, we use the analytic forward-noise strategy used in~\cite{OpenWAM}, and provide the detailed formulation in the appendix.

\subsection{Training Protocol}

We organize training into three stages according to the supervision available in our egocentric data pyramid. We use \emph{pre-training}, \emph{mid-training}, and \emph{post-training} to denote world-model pre-training on egocentric video data, joint world-action mid-training on action-annotated data, and downstream robot adaptation, respectively. During pre-training, we optimize only the world-modeling objective on egocentric video. Mid-training uses samples with reliable actions and jointly optimizes world and action prediction. Post-training adapts the model to downstream robot manipulation using embodiment-specific robot demonstrations, optionally together with robot-aligned human data. We keep the model architecture fixed across these stages unless otherwise specified. The appendix provides detailed data mixtures, optimization schedules, and implementation settings.

\section{Experiments}
\label{sec:experiments}

Our experiments follow the egocentric data pyramid, progressively relaxing constraints on human-data scaling. We begin with robot-aligned demonstrations to study OOD generalization and robustness. We then move to large-scale multi-task video-action data to examine transferable manipulation priors and scaling along data duration and task diversity. Finally, we relax action supervision to investigate whether video-only experience can further benefit robot learning.

We evaluate the effect of egocentric data on downstream robot performance through closed-loop evaluation, rather than relying on human-action prediction MSE as an open-loop proxy. For real-world evaluation, we use a bimanual Piper robot with parallel grippers and collect robot demonstrations through teleoperation. The appendix provides detailed task definitions, data collection protocols, and ID and OOD evaluation settings. For broader evaluation, we follow the official RoboDojo~\citep{RoboDojo} protocol and report both progress score and success rate. Since our model does not explicitly target long-term memory or VLM-based open instruction following, we focus primarily on \emph{Generalization}, \emph{Precision}, and \emph{Long-Horizon}, while reporting results across all benchmark categories for completeness.

\subsection{Robot-Aligned Human Demonstrations}
\label{sec:aligned_human}

We first study robot-aligned human demonstrations to examine two questions: whether they improve out-of-distribution (OOD) generalization and robustness to unseen object and scene variations, and whether they reduce the amount of target-task robot supervision required. In particular, we conduct real-robot experiments on \emph{Place into the Basket} and \emph{Fold Cloth}, representing rigid-object pick-and-place and deformable-object manipulation, respectively. For each task, we collect 300 in-domain (ID) robot demonstrations and a matched human set containing 300 ID, 100 object-OOD, and 100 scene-OOD demonstrations under the same task specification. The OOD conditions are absent from robot training and are covered only by the corresponding human demonstrations.

\begin{figure}[t]
\begin{center}
\includegraphics[width=\linewidth]{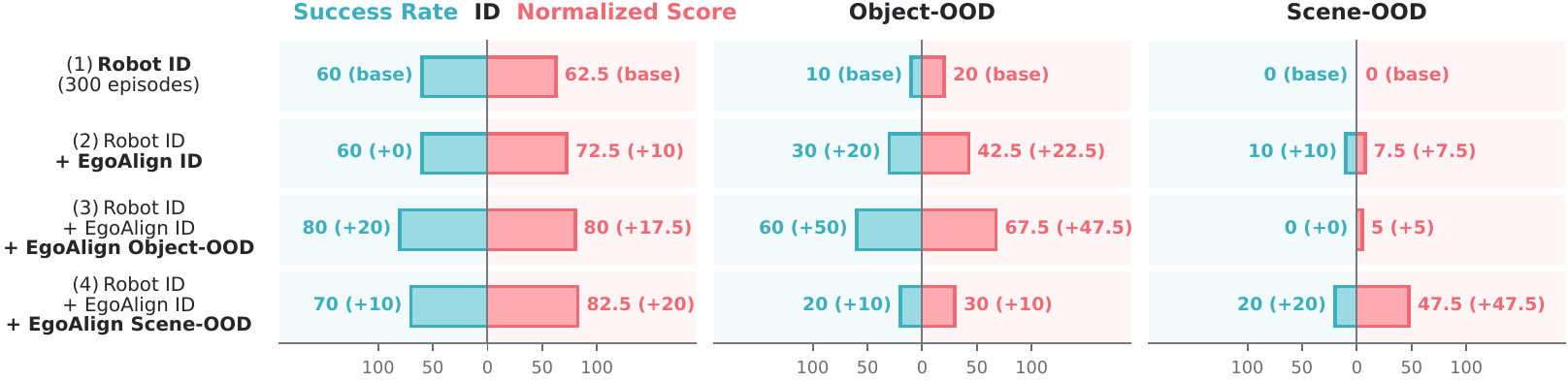}
\end{center}
\vspace{-1em}
\caption{Real-robot evaluation on \emph{Place into the Basket} with human data under OOD settings.}
\label{figure:posttrain_place}
\end{figure}

\begin{figure}[t]
\begin{center}
\includegraphics[width=\linewidth]{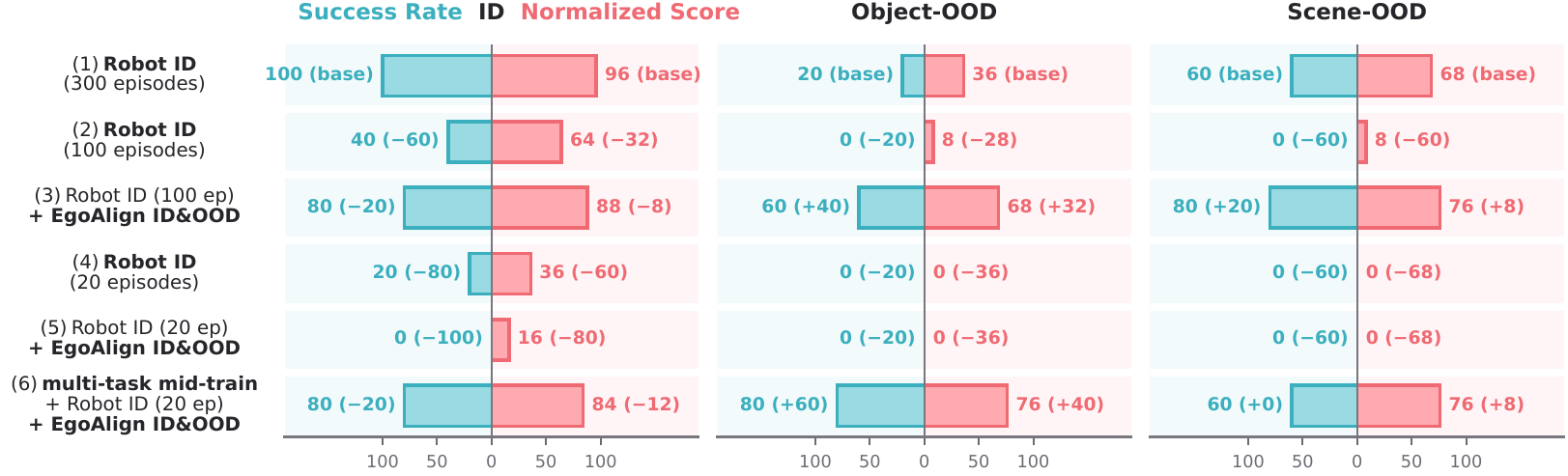}
\end{center}
\vspace{-1em}
\caption{Real-robot evaluation on \emph{Fold Cloth} with human data under different robot-data budgets.}
\label{figure:posttrain_fold_cloth}
\end{figure}

\vspace{0.05in}
\noindent\textbf{Generalizing to Environments Unseen in Robot Data.}
Figures~\ref{figure:posttrain_place} and~\ref{figure:posttrain_fold_cloth} show that human demonstrations from object or scene conditions absent from robot training can substantially improve robot performance under the corresponding conditions. On \emph{Place into the Basket}, the robot-only policy achieves 60\% ID success but drops to 10\% on object-OOD and 0\% on scene-OOD. Introducing object-OOD human demonstrations raises object-OOD success to 60\%, and introducing scene-OOD human demonstrations raises scene-OOD success to 20\% with a substantially larger gain in task progress. These results show that the model can acquire task-relevant knowledge about new objects and environments directly from human demonstrations. This provides a practical way to extend a robot policy to new objects and scenes without requiring corresponding robot demonstrations.


\vspace{0.05in}
\noindent\textbf{Reducing Dependence on Robot Data.}
We next examine whether aligned human demonstrations can reduce the amount of target-task robot data required for policy learning. We vary the number of robot demonstrations on \emph{Fold Cloth} while keeping the same 500 human demonstrations fixed. Reducing robot data from 300 to 100 demonstrations causes a substantial performance drop, whereas adding human data recovers 80\% ID, 60\% object-OOD, and 80\% scene-OOD success. However, when robot supervision is further reduced to only 20 demonstrations, the same human data is no longer sufficient to recover effective policy performance. These results show that aligned human demonstrations can substantially reduce the dependence on target-task robot data, but cannot fully replace embodiment-specific robot supervision.

\subsection{Scaling Beyond Task-Specific Egocentric Data}
\label{sec:midtrain}

Although robot-aligned human demonstrations can effectively help models adapt to specific distributions, their dependence on target-task collection limits scalability. We therefore move to large-scale multi-task egocentric data collected independently of downstream tasks and investigate whether this experience can establish transferable manipulation priors before downstream adaptation. We further study how scaling along two key dimensions, data duration and task diversity, affects downstream robot learning. We call this stage \emph{multi-task mid-training}, where we optimize both video and action objectives.
In addition to egocentric human data, we include approximately 90 hours of real-robot data to provide embodiment-specific grounding in the robot action space. The robot data contains only two generic task families, pick-and-place and towel folding, and the corresponding tasks, objects, and scenes do not overlap with downstream evaluation.

\vspace{0.05in}
\noindent\textbf{Further Reducing Dependence on Robot Data.}
As shown in Figure~\ref{figure:posttrain_fold_cloth}, with only 20 target-task robot demonstrations, adding the full aligned human set alone fails to recover useful performance. In contrast, after multi-task mid-training, the model achieves 80\% success on ID, 80\% on object-OOD, and 60\% on scene-OOD using the same 20 robot demonstrations and aligned human data. Importantly, the amount of target-task robot data and the aligned human data do not change between these settings. The improvement therefore reflects the transferable priors acquired during multi-task mid-training. This suggests that broad multi-task experience can make limited downstream data more effective, further reducing dependence on target-task robot data.

\begin{figure}[t]
\begin{center}
\includegraphics[width=\linewidth]{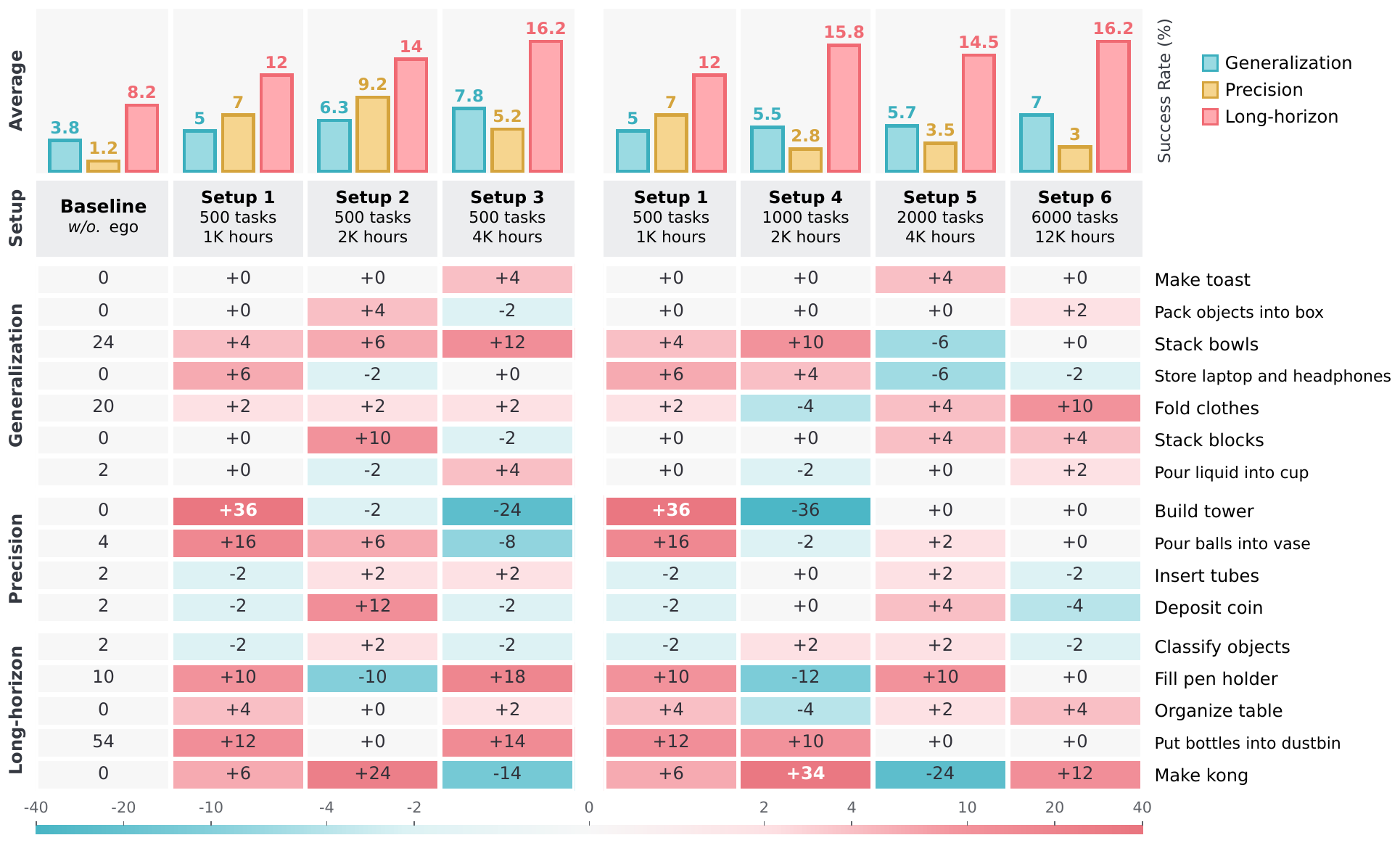}
\end{center}
\vspace{-1em}
\caption{\textbf{RoboDojo evaluation under data-duration and task-diversity scaling.} We visualize a subset of tasks with non-zero success rate, while all aggregate results are computed over the full evaluation task set. We provide complete per-task results in the appendix.}
\label{figure:midtrain_robodojo}
\end{figure}

\vspace{0.05in}
\noindent\textbf{Scaling Data Duration and Task Diversity.}
We next study how to scale multi-task human data by decomposing data scale into two dimensions: the data duration within a fixed task set and the diversity of tasks covered. We construct two controlled scaling sequences from the semantic task categories of our egocentric data. For \emph{duration scaling}, we fix the same 500 most frequent tasks and increase the total duration from 1K to 2K and 4K hours. For \emph{task scaling}, we progressively expand from 500 to 1K, 2K, and 6K tasks toward the naturally occurring long tail, while keeping the average duration per task approximately constant, resulting in 1K, 2K, 4K, and 12K hours of data.

As shown in Fig.~\ref{figure:midtrain_robodojo}, the two scaling directions lead to different capability profiles. With a fixed task set, increasing duration from 1K to 4K hours improves Generalization from 5.0 to 7.8 and Long-Horizon from 12.0 to 16.2. Increasing task diversity from 500 to 6K tasks similarly improves Generalization from 5.0 to 7.0 and Long-Horizon from 12.0 to 16.2, but Precision decreases as we introduce increasingly long-tail tasks. At matched data volumes of 2K and 4K hours, denser coverage of the top 500 tasks also remains stronger on several aggregate metrics than broader task sets. The task-level results further reveal substantial positive and negative transfer.

These results show that data duration and task diversity are distinct, capability-dependent scaling axes. More data within existing tasks and broader task coverage benefit different capabilities and can also introduce trade-offs. Consequently, neither total hours nor task count alone adequately characterizes useful egocentric-data scale. The results suggest ensuring sufficient coverage of common behaviors before expanding aggressively into increasingly sparse long-tail tasks.

\subsection{Relaxing Action Supervision with Large-Scale Human Video}
\label{sec:video_pretrain}

As egocentric data scales, obtaining reliable action supervision becomes increasingly costly. Human action labels rely on accurate hand reconstruction, camera tracking, and temporal quality control, while reconstruction failures and noisy trajectories further reduce usable video-action data. In contrast, the corresponding videos can still preserve informative object motion, interaction outcomes, and physical state transitions. Since WAMs can learn such dynamics directly through video prediction, we investigate whether video-only supervision can improve downstream robot learning without requiring reliable action labels, and how its benefits interact with subsequent video-action mid-training.

We call this stage \emph{video pre-training}, where we optimize only the video-prediction objective. The pre-training set contains the same 12K hours of action-valid egocentric videos used for mid-training, plus approximately 3K additional hours sampled from the video-only data, for a total of 15K hours. Mid-training subsequently uses the 12K-hour action-valid subset together with the multi-task robot data described in Section~\ref{sec:midtrain}, jointly optimizing video and action objectives. The two stages are not designed as a compute-matched comparison, since video-only data is naturally available at a larger scale than reliably action-annotated data. This scale difference is reflected in the egocentric data pyramid and motivates studying how the two supervision regimes can be used together.

\begin{table*}[t]
\centering

\caption{
\textbf{Ablation study of egocentric data utilization strategies on RoboDojo.}
Base and joint denote different shared-attention mechanisms used during post-training.
}

\label{table:robodojo_ablation}

\vspace{0.5em}

\resizebox{\linewidth}{!}{
\setlength{\tabcolsep}{1.2mm}{

\begin{tabular}{ccc|cccccccccc|cc}

\toprule

\multicolumn{3}{c|}{\textbf{Training Setup}} &
\multicolumn{2}{c}{\textbf{{Genera.}}} &
\multicolumn{2}{c}{\textbf{Precision}} &
\multicolumn{2}{c}{\textbf{{Long}}} &
\multicolumn{2}{c}{\textbf{Memory}} &
\multicolumn{2}{c|}{\textbf{Open}} &
\multicolumn{2}{c}{\textbf{Average}} \\

\cmidrule(lr){1-3}
\cmidrule(lr){4-5}
\cmidrule(lr){6-7}
\cmidrule(lr){8-9}
\cmidrule(lr){10-11}
\cmidrule(lr){12-13}
\cmidrule(lr){14-15}

pre-train &
mid-train &
post-train &
Score &
SR &
Score &
SR &
Score &
SR &
Score &
SR &
Score &
SR &
Score &
SR \\

\midrule

- & - & base &
5.87 & 3.83 &
5.56 & 1.25 &
16.97 & 8.25 &
2.75 & 1.67 &
0.78 & 0.75 &
6.39 & 3.15 \\

- & $\checkmark$ & base &
10.73 & 7.00 &
8.45 & 3.00 &
25.18 & 16.25 &
6.38 & 5.33 &
0.68 & 0.50 &
10.28 & 6.42 \\

$\checkmark$ & - & base &
12.04 & 7.67 &
23.53 & 16.25 &
27.85 & 17.50 &
6.67 & 5.33 &
0.55 & 0.50 &
14.13 & 9.45 \\

$\checkmark$ & $\checkmark$ & base &
12.63 & 8.33 &
24.77 & 18.00 &
29.46 & 20.50 &
7.42 & 6.33 &
0.90 & 0.75 &
15.04 & 10.78 \\

$\checkmark$ & - & joint &
17.19 & 12.17 &
31.16 & 22.25 &
44.16 & 29.75 &
8.37 & 7.33 &
0.55 & 0.50 &
20.29 & 14.35 \\

\bottomrule

\end{tabular}
}}

\end{table*}
\begin{table*}[t]
\centering

\caption{
\textbf{Performance comparison on RoboDojo.}
The scale of robot and egocentric training data is reported in hours.
Results of existing methods are taken from the official leaderboard.
The style of \textbf{bold} and \underline{underline} denotes the best and second-best results, respectively.
}

\label{table:robodojo_comparison}

\vspace{0.5em}

\resizebox{\linewidth}{!}{
\setlength{\tabcolsep}{1.2mm}{

\begin{tabular}{l|cc|cccccccccc|cc}

\toprule

\multirow{2}{*}{\textbf{Model}} &
\multicolumn{2}{c|}{\textbf{Data Scale}} &
\multicolumn{2}{c}{\textbf{{Genera.}}} &
\multicolumn{2}{c}{\textbf{Precision}} &
\multicolumn{2}{c}{\textbf{{Long}}} &
\multicolumn{2}{c}{\textbf{Memory}} &
\multicolumn{2}{c|}{\textbf{Open}} &
\multicolumn{2}{c}{\textbf{Average}} \\

\cmidrule(lr){2-3}
\cmidrule(lr){4-5}
\cmidrule(lr){6-7}
\cmidrule(lr){8-9}
\cmidrule(lr){10-11}
\cmidrule(lr){12-13}
\cmidrule(lr){14-15}

&
Robot &
Ego &
Score &
SR &
Score &
SR &
Score &
SR &
Score &
SR &
Score &
SR &
Score &
SR \\

\midrule

Pi-0.5 &
- & - &
13.38 & 8.17 &
12.40 & 5.50 &
23.54 & 14.67 &
5.89 & 4.67 &
1.98 & 1.67 &
11.44 & 6.93 \\

Spatial Forcing &
- & - &
14.12 & 9.34 &
17.32 & 10.58 &
23.26 & 14.58 &
5.43 & 4.11 &
1.78 & 1.58 &
12.38 & 8.04 \\

Hy-0.5-VLA &
- & - &
11.78 & 8.39 &
13.81 & 8.00 &
25.74 & 14.92 &
\underline{13.37} & \underline{12.11} &
0.65 & 0.58 &
13.07 & 8.80 \\

Meituan-0 &
- & - &
13.75 & 8.17 &
16.77 & 7.75 &
29.61 & 18.58 &
10.06 & 8.89 &
\underline{4.54} & \underline{4.25} &
14.95 & 9.53 \\

Xiaomi-1 &
7.2K & 100K &
\underline{23.54} & \underline{17.00} &
26.69 & 18.83 &
38.39 & 23.67 &
7.81 & 6.56 &
3.94 & 3.58 &
20.07 & 13.93 \\

GalaxeaVLA-0.5 &
- & - &
18.46 & 12.83 &
\underline{28.25} & \underline{20.42} &
\underline{44.12} & \textbf{32.25} &
8.61 & 7.33 &
1.73 & 1.58 &
20.23 & \underline{14.88} \\

GPT-6-Astra &
- & - &
\textbf{33.36} & \textbf{30.50} &
12.65 & 4.00 &
21.45 & 8.25 &
\textbf{43.04} & \textbf{38.67} &
\textbf{34.36} & \textbf{31.00} &
\textbf{28.97} & \textbf{22.48} \\

\midrule

FastWAM$^\star$ &
\multicolumn{2}{c|}{\small{\textit{w/o.} pretrain}} &
5.87 & 3.83 &
5.56 & 1.25 &
16.97 & 8.25 &
2.75 & 1.67 &
0.78 & 0.75 &
6.39 & 3.15 \\

OpenWAM-$\alpha$ &
4.9K & 1.4K &
20.71 & 14.83 &
18.45 & 9.25 &
34.93 & 25.33 &
10.41 & 9.11 &
1.41 & 1.08 &
17.18 & 11.92 \\

\textbf{Ego4WAM} &
0.1K & 15K &
17.19 & 12.17 &
\textbf{31.16} & \textbf{22.25} &
\textbf{44.16} & \underline{29.75} &
8.37 & 7.33 &
0.55 & 0.50 &
\underline{20.29} & 14.35 \\

\bottomrule

\end{tabular}
}}

\end{table*}

\vspace{0.05in}
\noindent\textbf{Learning Dynamics Priors without Action Supervision.}
As shown in Table~\ref{table:robodojo_ablation}, the model without human pre-training or mid-training achieves an average RoboDojo score of 6.39 with a 3.15\% success rate. Video-only pre-training improves this to 14.13 / 9.45\% despite using no human action supervision, showing that egocentric experience remains valuable even when reliable action labels are unavailable. WAMs can therefore extend the usable data beyond the subset that supports imitation learning.

Applying video-action mid-training after pre-training further improves performance from 14.13 / 9.45\% to 15.04 / 10.78\%. This relatively small gain suggests that the two stages capture partially overlapping information about manipulation dynamics. More importantly, the same mid-training without prior video pre-training reaches only 10.28 / 6.42\%. The substantial gap between these two settings highlights the importance of first establishing a strong world-modeling branch for WAMs, so that subsequent action learning can build on informative visual dynamics. Egocentric video is particularly suitable for this role, as it can be scaled without action labels while retaining dynamics.

\vspace{0.05in}
\noindent\textbf{Leveraging Dynamics Priors for Joint World-Action Learning.}
The results above show that large-scale video pre-training provides a strong dynamics prior for the world-modeling branch. We next ask whether this improved world representation can be more effectively exploited during downstream action learning. We compare the \emph{base} and \emph{joint} shared-attention mechanisms introduced in Section~\ref{sec:wam_framework}, where the joint formulation additionally allows action prediction to access predicted future visual representations. Starting from the same video-pretrained model, switching from base to joint substantially improves the average RoboDojo score from 14.13 to 20.29 and the success rate from 9.45\% to 14.35\%. This suggests that stronger dynamics priors make predicted future representations more informative for action generation, allowing joint world-action learning to better exploit the knowledge acquired from egocentric video.

Interestingly, FastWAM~\citep{FastWAM} reports only small differences between its base and joint variants without embodied pre-training. Our results suggest a complementary explanation that the benefit of joint future conditioning depends on the quality of the learned world representation. With large-scale egocentric video pre-training, the same basic WAM backbone achieves an average RoboDojo score of 20.29 and a 14.35\% success rate using only approximately 0.1K hours of multi-task robot data and 15K hours of egocentric experience. This result further demonstrates the potential of egocentric data to strengthen world dynamics in WAMs and highlights the importance of studying not only how much human data is available, but also how it is used throughout the training pipeline.

\subsection{More Results on the Real Robot Tasks}

\begin{figure}[t]
\begin{center}
\includegraphics[width=\linewidth]{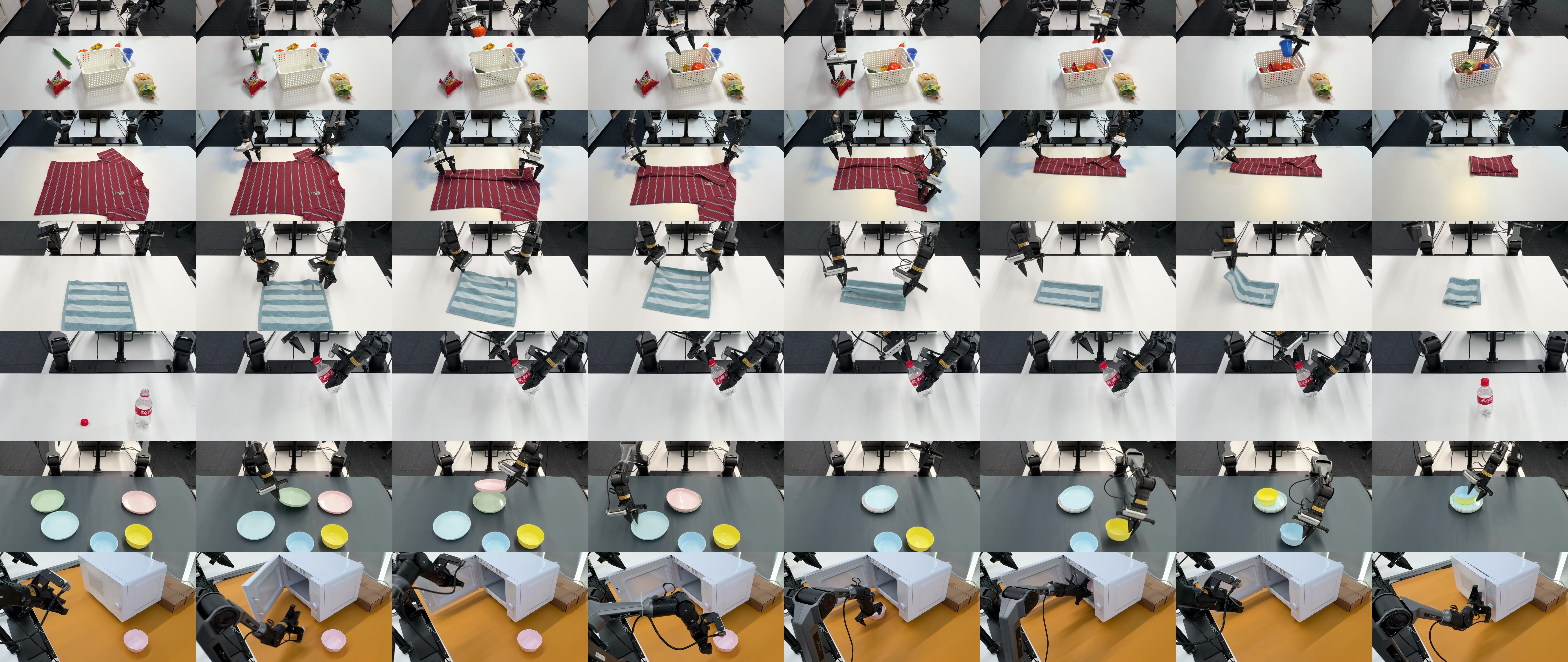}
\end{center}
\vspace{-1em}
\caption{Real-robot evaluation across diverse manipulation tasks.}
\label{figure:real_robot}
\end{figure}

Finally, we evaluate on a broader set of real-world manipulation tasks beyond the settings used above. As shown in Figure~\ref{figure:real_robot}, we consider tasks spanning rigid-object manipulation, deformable-object interaction, and longer-horizon composition: \emph{Fold Towel}, \emph{Tighten a Bottle Cap}, \emph{Stack Plates and Bowls}, and \emph{Heat Food with a Microwave}. We provide videos on the project page.

\section{Conclusion}
\label{sec:conclusion}

We presented a systematic study on scaling egocentric human data for robot learning with WAM. Rather than treating the duration of human data as a single measure of scale, we disentangled the roles of human-robot alignment, data duration and task diversity, available supervision, and usage strategies. The study leads to several key \textbf{takeaways} for scaling egocentric human data:

\begin{itemize}[leftmargin=*]
    \item \textbf{Human data reduces the need for robot data, but cannot fully replace it.}
    Broad multi-task human experience provides transferable manipulation priors, aligned human demonstrations cover distribution variations, and a small amount of robot data remains important for embodiment adaptation. Their combination can substantially reduce the amount of target-task robot data.

    \item \textbf{Scaling egocentric data does not always lead to better performance.}
    Data duration and task diversity represent different ways to scale, and total hours alone cannot characterize useful data scale. Denser coverage of recurring tasks provides limited additional gains, while introducing increasingly sparse long-tail tasks can even hurt precision-sensitive performance.

    \item \textbf{Egocentric data provides a strong dynamics prior for world-action learning.}
    Video-only pre-training can learn strong dynamics priors without action labels, extending usable data beyond the action subset. Such priors further improve subsequent world-action learning, showing that the value of human data depends not only on its supervision, but also on when and how it is used.
\end{itemize}

Ego4WAM provides a practical perspective on how to collect and utilize human data for robot learning. We hope this study serves as an empirical reference for designing future egocentric-data pipelines, and we will release our code, training recipes, and model checkpoints at different training stages to support further research.

\section*{Statement}

\vspace{0.05in}
\noindent\textbf{AI Use Statement.} Generative AI tools were used to assist with language editing, manuscript organization, and improving the clarity of presentation. The authors developed, reviewed, and verified all research ideas, experimental design, data processing, analyses, and conclusions. The authors take full responsibility for the content of this paper.

\vspace{0.05in}
\noindent\textbf{Ethics Statement.} Our study involves egocentric recordings of human manipulation collected for research purposes. All participants in our internally collected data provided informed consent for data collection and research use. For publicly available datasets, we follow their respective licenses, terms of use, and data-use requirements. We do not intentionally collect or release personally identifiable information, and we use the collected data solely for research on robot learning.

\vspace{0.05in}
\noindent\textbf{Reproducibility Statement.} We provide detailed descriptions of the model framework, action space, training stages, and evaluation protocols in the main paper and appendix. To facilitate reproducibility and further research, we will release our code, training recipes, and model checkpoints from different pre-training and mid-training stages. We will also release the RoboDojo post-trained checkpoints and submit them to the official RoboDojo evaluation pipeline, allowing independent verification of our reported benchmark results and direct comparison under the standardized evaluation protocol.


\bibliography{iclr2027_conference}
\bibliographystyle{iclr2027_conference}

\newpage
\appendix

\section{Egocentric Data}
\label{app:ego_data}

\subsection{Egocentric Data Pyramid}
\label{app:data_pyramid}

Figure~\ref{figure:egocentric_data_pyramid} summarizes the supervision hierarchy and available scale of the egocentric data considered in our study. The pyramid characterizes a general property of egocentric data rather than the exact data mixture used in each experiment. As stronger supervision is required, fewer samples can reliably satisfy the corresponding annotation and alignment requirements. Robot-aligned demonstrations therefore occupy the smallest tier, followed by multi-task video-action data with reliable human actions, while video-only data can be retained at substantially larger scale.

We construct the pyramid based on the most reliable supervision each sample supports. We process EgoAlign separately as the robot-aligned subset. For public datasets with available hand or action annotations, we apply temporal and geometric quality-control filters to remove unreliable trajectories. Samples that pass the action-quality checks form the multi-task video-action data. Samples without reliable action annotations, including those rejected by the action-quality pipeline, can still be retained as video-only data when the underlying visual interaction remains valid. This way, unreliable action labels do not require discarding otherwise useful interaction videos.

Importantly, the approximately 120K hours of video-only data shown in Figure~\ref{figure:egocentric_data_pyramid} represent the available data scale and are not all used in our experiments. Due to practical training constraints, we construct task-specific training subsets from this collection. In our main video pre-training experiments, we use the 12K-hour action-valid subset with video supervision only, plus about 3K additional hours sampled from the video-only data, for a total of 15K hours of egocentric video. We then use the 12K-hour action-valid subset for multi-task mid-training with both video and action supervision. Thus, the pyramid primarily shows how the amount of usable egocentric experience increases as supervision requirements are relaxed, while the corresponding experiments specify the exact data mixtures used at each training stage.

\subsection{EgoAlign}
\label{app:egoalign}

EgoAlign is our internally collected egocentric dataset for studying robot-aligned human demonstrations. We record human demonstrations using a monocular head-mounted camera under task and environment settings matching the real-robot experiments. All participants provided informed consent for data collection and research use. We recover human hand actions from the recorded videos, referring to the general processing pipeline of OpenAoE~\citep{OpenAoE}.

Specifically, we first estimate the camera trajectory using VGGT-Omega~\citep{VGGTOmega} and take the coordinate system of the first camera frame as the reference frame for each sequence. We then use GroundingDINO~\citep{GroundingDINO} to localize hand regions and apply HaMeR~\citep{HaMeR} to estimate the 3D hand poses independently for each frame. Combining the estimated hand pose with the recovered camera trajectory transforms the frame-wise hand predictions into a temporally shared coordinate system referenced to the first camera frame. The resulting trajectories provide the hand motion used to construct the shared end-effector and gripper supervision for human demonstrations.
Finally, we inspect and filter the reconstructed trajectories to remove samples with unreliable hand localization, implausible hand motion, unstable camera tracking, or obvious temporal and geometric inconsistencies. We retain only sequences that pass this quality-control process as action-annotated EgoAlign demonstrations.

\section{Method}

\subsection{Unified Action Space}
\label{app:action_space}

\vspace{0.05in}
\noindent\textbf{Converting Human Hand Pose to Gripper EEF.}
To jointly train on human and robot actions, we abstract each human hand as a parallel-gripper end effector. Each hand is represented by a 9D end-effector pose together with a 1D gripper-openness signal. The 3D end-effector position is defined as the midpoint between the thumb and index fingertips,
\begin{equation}
    \mathbf{p}_{\mathrm{eef}}
    =
    \frac{1}{2}
    \left(
    \mathbf{p}_{\mathrm{thumb}}
    +
    \mathbf{p}_{\mathrm{index}}
    \right),
\end{equation}
while gripper openness is defined by their Euclidean distance,
\begin{equation}
    g =
    \left\|
    \mathbf{p}_{\mathrm{thumb}}
    -
    \mathbf{p}_{\mathrm{index}}
    \right\|_2 .
\end{equation}

For hand orientation, we follow the hand-centric construction of ACE-Ego-0~\citep{ACEEgo0}. Let $\mathbf{p}_{\mathrm{wrist}}$ denote the wrist position and let
$\mathbf{p}_{\mathrm{palm}}$ be the centroid of the index, middle, and ring fingertips. We construct an orthonormal hand frame as
\begin{equation}
    \mathbf{x}
    =
    \frac{
        \mathbf{p}_{\mathrm{palm}}-\mathbf{p}_{\mathrm{wrist}}
    }{
        \left\|
        \mathbf{p}_{\mathrm{palm}}-\mathbf{p}_{\mathrm{wrist}}
        \right\|_2
    },
    \qquad
    \mathbf{z}
    =
    \hat{\mathbf{n}}
    \left(
        \mathbf{p}_{\mathrm{wrist}},
        \mathbf{p}_{\mathrm{thumb}},
        \mathbf{p}_{\mathrm{middle}}
    \right),
    \qquad
    \mathbf{y} = \mathbf{z}\times\mathbf{x},
\end{equation}
where $\hat{\mathbf{n}}(\cdot)$ denotes the unit normal of the corresponding hand plane, with its direction chosen consistently with the palm orientation. The resulting rotation matrix
$\mathbf{R}_{\mathrm{hand}}=[\mathbf{x},\mathbf{y},\mathbf{z}]$
is converted to the continuous 6D rotation representation by concatenating its first two columns. Together with the 3D position, this yields the 9D EEF representation used for human action supervision.

Robot actions additionally contain embodiment-specific joint signals. As described in the main paper, each robot arm is represented by 6D joint states, a 1D gripper state, and the same 9D EEF representation. Human demonstrations supervise only the shared EEF and gripper components, leaving robot-specific joint dimensions unavailable.

\vspace{0.05in}
\noindent\textbf{Analytic Noise Path.}
Following the unified action interface of OpenWAM~\citep{OpenWAM}, we embed embodiment-specific actions into a fixed-width action space and associate each sample with a validity mask indicating which action dimensions are supervised. Robot demonstrations provide both joint and shared EEF supervision, whereas human demonstrations activate only the shared EEF and gripper dimensions. We denote the active and inactive dimension sets by $\mathcal{A}$ and $\mathcal{I}$, respectively.

We train the action branch with flow matching. Given a clean action $\mathbf{x}_0$, Gaussian noise $\boldsymbol{\epsilon}\sim\mathcal{N}(0,I)$, and noise level $\sigma$, the forward trajectory is
\begin{equation}
    \mathbf{x}_{\sigma}
    =
    (1-\sigma)\mathbf{x}_0
    +
    \sigma\boldsymbol{\epsilon},
    \qquad
    \mathbf{v}^{\star}
    =
    \boldsymbol{\epsilon}-\mathbf{x}_0 .
\end{equation}
Only valid action dimensions contribute to the training objective,
\begin{equation}
    \mathcal{L}_{\mathrm{action}}
    =
    \mathbb{E}
    \left[
    w(\sigma)
    \frac{
        \sum_{t,d} M_{t,d}
        \left\|
        \mathbf{v}_{\theta}(\mathbf{x}_{\sigma},\sigma)_{t,d}
        -
        \mathbf{v}^{\star}_{t,d}
        \right\|_2^2
    }{
        \max\left(\sum_{t,d}M_{t,d},1\right)
    }
    \right],
\end{equation}
where $M_{t,d}$ denotes the validity mask over temporal and action dimensions.

Simply masking the loss leaves predictions on inactive dimensions unconstrained, which can accumulate arbitrary values during iterative flow sampling. We therefore keep inactive dimensions on their analytic forward-noise trajectory. Since $\mathbf{x}_0[\mathcal{I}]=0$, their training marginal is
$\mathbf{x}_{\sigma}[\mathcal{I}]=\sigma\boldsymbol{\epsilon}_{\mathcal{I}}$.
At the first sampling step, we recover the corresponding latent noise and keep it fixed throughout sampling,
\begin{equation}
    \hat{\boldsymbol{\epsilon}}_{\mathcal{I}}
    =
    \frac{
        \mathbf{x}_{\sigma_0}[\mathcal{I}]
    }{\sigma_0},
    \qquad
    \mathbf{x}_{\sigma_i}[\mathcal{I}]
    =
    \sigma_i
    \hat{\boldsymbol{\epsilon}}_{\mathcal{I}} .
\end{equation}
Thus, inactive dimensions remain distributed consistently with the training forward process without contributing supervision or being numerically integrated by the sampler.

\subsection{Optimization Objectives}
\label{app:optimization}

Following FastWAM~\citep{FastWAM}, we train both the world-modeling and action-prediction branches with conditional flow matching. Let $\mathbf{y}$ denote either future video latents or an action trajectory. Given Gaussian noise
$\boldsymbol{\epsilon}\sim\mathcal{N}(0,I)$ and a noise level
$\sigma\in(0,1)$, we construct the noisy sample as
\begin{equation}
    \mathbf{y}_{\sigma}
    =
    (1-\sigma)\mathbf{y}
    +
    \sigma\boldsymbol{\epsilon},
\end{equation}
with the corresponding velocity target
\begin{equation}
    \mathbf{v}^{\star}
    =
    \boldsymbol{\epsilon}-\mathbf{y}.
\end{equation}
The generic flow-matching objective is
\begin{equation}
    \mathcal{L}_{\mathrm{FM}}(\mathbf{y})
    =
    \mathbb{E}_{\mathbf{y},\boldsymbol{\epsilon},\sigma}
    \left[
        w(\sigma)
        \left\|
        \mathbf{v}_{\theta}
        (\mathbf{y}_{\sigma},\sigma,\mathbf{c})
        -
        \mathbf{v}^{\star}
        \right\|_2^2
    \right],
\end{equation}
where $\mathbf{c}$ denotes the corresponding conditioning signals and
$w(\sigma)$ is the timestep-dependent loss weight.

\noindent\textbf{World-Modeling Objective.}
For world modeling, $\mathbf{y}=\mathbf{z}_{1:T}$ denotes the latent representation of future video frames encoded by the video VAE. The world-modeling loss is
\begin{equation}
    \mathcal{L}_{\mathrm{world}}
    =
    \mathcal{L}_{\mathrm{FM}}(\mathbf{z}_{1:T}).
\end{equation}
The world branch is conditioned on the current visual observation and language instruction. This objective applies to egocentric videos regardless of whether reliable action annotations are available.

\noindent\textbf{Action-Prediction Objective.}
For action prediction, $\mathbf{y}=\mathbf{a}_{1:H}$ denotes an action chunk of horizon $H$. We optimize
\begin{equation}
    \mathcal{L}_{\mathrm{action}}
    =
    \mathcal{L}_{\mathrm{FM}}(\mathbf{a}_{1:H}),
\end{equation}
where the action branch is additionally conditioned on the available proprioceptive state. For human demonstrations, we evaluate the loss only over the shared EEF and gripper dimensions defined in Section~\ref{app:action_space}; we handle unavailable embodiment-specific dimensions using the validity mask and analytic noise path described above.

\noindent\textbf{Training with Different Supervision.}
The overall objective for an action-annotated sample is
\begin{equation}
    \mathcal{L}
    =
    \lambda_{\mathrm{world}}\mathcal{L}_{\mathrm{world}}
    +
    \lambda_{\mathrm{action}}\mathcal{L}_{\mathrm{action}},
\end{equation}
where $\lambda_{\mathrm{world}}$ and $\lambda_{\mathrm{action}}$ control the relative contribution of the two objectives. For samples trained without action supervision, we optimize only
$\mathcal{L}_{\mathrm{world}}$.

This formulation allows the same WAM backbone to naturally accommodate the different supervision levels in our egocentric data pyramid. During video pre-training, we optimize all samples only with $\mathcal{L}_{\mathrm{world}}$, including both action-valid videos and additional video-only data. During multi-task mid-training, we optimize action-annotated human and robot samples jointly with $\mathcal{L}_{\mathrm{world}}$ and $\mathcal{L}_{\mathrm{action}}$. Post-training follows the same joint objective on downstream robot demonstrations, optionally together with aligned human demonstrations.

\subsection{Training Details}
\label{app:training_details}

Our implementation is built upon the StarVLA codebase~\citep{StarVLA}, with modifications to support our WAM architecture, unified human-robot action space, and multi-stage training pipeline.

We use five training configurations across our experiments: video pre-training, full-scale multi-task mid-training, scaling-study mid-training, real-robot post-training, and RoboDojo post-training. Table~\ref{tab:training_details} summarizes the detailed optimization settings. Unless otherwise specified, all experiments use a per-GPU batch size of 16 with 8 NVIDIA H20 GPUs per node, cosine learning-rate decay, 8-frame future video prediction, and a 32-step action prediction horizon.

For the scaling study, all data-duration and task-diversity variants use the same training configuration and number of optimization steps; only the scale and composition of the egocentric training data are changed. Similarly, all real-robot tasks and robot-data-budget ablations share the same post-training recipe, and all RoboDojo experiments use the same RoboDojo post-training configuration.

\begin{table*}[t]
\centering
\caption{
\textbf{Training configurations used throughout our experiments.}
All experiments within the same configuration use identical optimization settings unless otherwise specified.
}
\label{tab:training_details}

\vspace{0.5em}

\resizebox{\linewidth}{!}{
\setlength{\tabcolsep}{1mm}{
\begin{tabular}{lccccc}
\toprule
Parameter
& Video Pre-train
& Full Mid-train
& Scaling Mid-train
& Real-Robot Post-train
& RoboDojo Post-train \\
\midrule

Training Steps
& 150K
& 80K
& 40K
& 10K
& 40K \\

Nodes
& 8
& 8
& 4
& 1
& 2 \\

Initial LR
& 2e-5
& 1e-4
& 1e-4
& 1e-4
& 1e-4 \\

Final LR
& 1e-5
& 1e-6
& 1e-6
& 1e-6
& 1e-6 \\

Warmup Steps
& 2K
& 2K
& 2K
& 1K
& 2K \\

Shared Attention
& Base
& Base
& Base
& Base
& Base / Joint \\

\bottomrule
\end{tabular}
}}
\end{table*}

\section{Experiments}
\label{app:experiments}

\subsection{Real-Robot Setup}
\label{app:real_robot}

\vspace{0.05in}
\noindent\textbf{Robot Platform and Data Collection.}
We conduct real-world experiments using a bimanual Piper robot equipped with parallel grippers. Robot demonstrations are collected through teleoperation. We record human demonstrations in EgoAlign separately using a monocular head-mounted camera under the corresponding task and environment settings, then convert them into the shared gripper action representation described in Section~\ref{app:action_space}. Human and robot demonstrations therefore share the same task specification and environment conditions, while retaining their native viewpoints and embodiments.

\vspace{0.05in}
\noindent\textbf{Evaluation Tasks.}
Our controlled real-robot experiments focus on two representative manipulation tasks, \emph{Place into the Basket} and \emph{Fold Cloth}, covering rigid-object pick-and-place and deformable-object manipulation, respectively. For each task, we collect 300 in-domain (ID) robot demonstrations. The corresponding EgoAlign set contains 300 ID, 100 object-OOD, and 100 scene-OOD human demonstrations. The object- and scene-OOD conditions are absent from robot training unless otherwise specified, allowing us to evaluate whether human demonstrations can provide information about environment variations not covered by robot data.

Object-OOD evaluation varies the manipulated objects beyond those appearing in the ID robot demonstrations, including changes in object category, instance, appearance, or size. Scene-OOD evaluation changes the surrounding visual environment, including tabletop or tablecloth appearance and illumination conditions. These two axes capture complementary forms of distribution shift and need not be strictly mutually exclusive.

In addition to these controlled experiments, we evaluate the final model on a broader set of four real-world manipulation tasks: \emph{Fold Towel}, \emph{Tighten a Bottle Cap}, \emph{Stack Plates and Bowls}, and \emph{Heat Food with a Microwave}. These tasks span rigid-object manipulation, deformable-object interaction, and longer-horizon manipulation.

\vspace{0.05in}
\noindent\textbf{Evaluation Metrics.}
We evaluate all policies through closed-loop execution and report both task success rate and normalized task progress. Success rate is a binary metric that records whether the complete task is successfully accomplished within a rollout. Normalized task progress provides a finer-grained measure of partial task completion on a scale from 0 to 100. We define task-specific intermediate milestones according to the natural execution stages of each task and apply the same scoring criteria to all compared methods and evaluation settings.

For \emph{Place into the Basket}, each rollout contains three target objects. Successfully placing each object into the basket contributes 25 points. If all three objects are successfully grasped on their first grasp attempt, an additional 25 points are awarded. A rollout is considered successful when all three objects are placed into the basket.

For \emph{Fold Cloth}, the task is divided into four stages: completing the first fold, pulling the cloth into the required configuration, completing the second fold, and completing the third fold. Each completed stage contributes 20 points. If all four stages are completed successfully in a single execution without retries, an additional 20 points are awarded. A rollout is considered successful when the complete folding procedure is finished.

\begin{figure}[t]
\begin{center}
\includegraphics[width=\linewidth]{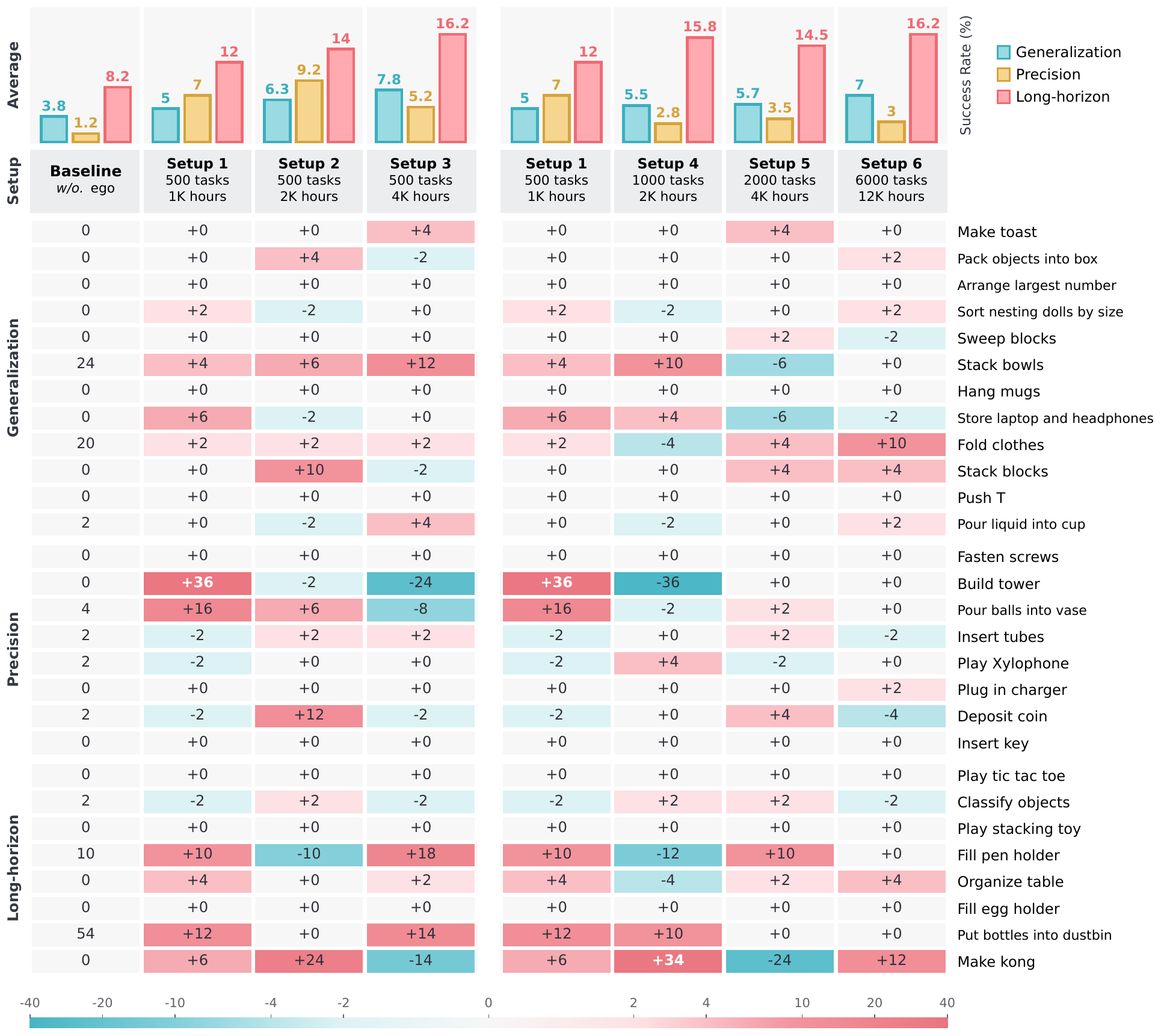}
\end{center}
\vspace{-1em}
\caption{RoboDojo evaluation under data-duration and task-diversity scaling.}
\label{appendix:figure:midtrain_robodojo_full}
\end{figure}

\subsection{RoboDojo Setup}
\label{app:robodojo}

We use RoboDojo~\citep{RoboDojo} as a standardized closed-loop benchmark for evaluating manipulation policies across a broad range of tasks. We follow the official RoboDojo post-training and evaluation protocol and report both progress score and success rate. All RoboDojo experiments use the same post-training configuration described in Section~\ref{app:training_details}, allowing comparison of models with different egocentric pre-training and mid-training strategies under a consistent downstream setting.

RoboDojo organizes its evaluation tasks into five categories: \emph{Generalization}, \emph{Precision}, \emph{Long-Horizon}, \emph{Memory}, and \emph{Open}. We report results on all five categories as well as the official average over the complete evaluation suite. Since our model does not explicitly introduce a long-term memory mechanism or a VLM-based module for open-ended instruction following, our analysis in the main paper focuses primarily on \emph{Generalization}, \emph{Precision}, and \emph{Long-Horizon}, while \emph{Memory} and \emph{Open} are retained for completeness.

For the data-scaling study, we compute all reported category averages over the complete RoboDojo evaluation task set. For readability, the task-level visualization in the main paper shows only tasks on which at least one evaluated setting achieves successful rollouts. The complete per-task results, including tasks with zero success across the compared settings, are provided in Figure~\ref{appendix:figure:midtrain_robodojo_full}.

\subsection{Data Duration and Task-Diversity Scaling}
\label{app:scaling_construction}

We construct the scaling subsets according to semantic task categories in the egocentric data. The task distribution is naturally imbalanced: a relatively small number of common manipulation behaviors account for a large fraction of the available data, while many less frequent tasks contain substantially fewer demonstrations. We refer to this less frequent portion of the empirical task distribution as the \emph{long tail}.

For duration scaling, we fix the 500 most frequent tasks and progressively increase the amount of data sampled from this fixed task set from 1K to 2K and 4K hours. This increases experience density within a common set of behaviors without expanding task coverage. For task-diversity scaling, we progressively expand the task set from the most frequent categories toward less frequent ones, using 500, 1K, 2K, and 6K tasks. We approximately preserve the average amount of data per task, resulting in 1K, 2K, 4K, and 12K hours of training data, respectively. As the task set grows, we introduce increasingly rare categories from the tail of the empirical task distribution into training.

All variants use the same model initialization, robot-data component, optimization schedule, and RoboDojo post-training protocol. The two scaling sequences therefore emphasize two different ways of increasing egocentric data scale: denser coverage of recurring behaviors and broader task coverage extending toward the long tail.

\end{document}